\documentclass{article}

\PassOptionsToPackage{numbers, compress}{natbib}

\usepackage[table]{xcolor} 
\usepackage[preprint]{neurips_2024}

\usepackage[utf8]{inputenc} % allow utf-8 input
\usepackage[T1]{fontenc}    % use 8-bit T1 fonts
\usepackage{hyperref}       % hyperlinks
\usepackage{url}            % simple URL typesetting
\usepackage{booktabs}       % professional-quality tables
\usepackage{amsfonts}       % blackboard math symbols
\usepackage{nicefrac}       % compact symbols for 1/2, etc.
\usepackage{microtype}      % microtypography
\usepackage{xcolor, graphicx} % colors
\usepackage{amsmath}
\usepackage{subcaption}
\usepackage{makecell}
\usepackage{algorithm}
\usepackage{algpseudocode}
\usepackage{multirow}
\usepackage{subcaption}

\title{Teaching with Uncertainty: Unleashing the Potential of Knowledge Distillation in Object Detection}

\author{
  Junfei Yi \\
   Hunan University, China\\
   \texttt{yijunfei@hnu.edu} \\
   \And
    Jianxu Mao \\
   Hunan University, China\\
   \texttt{maojianxu@hnu.edu} \\
   \And
   Tengfei Liu \\
   Beijing University of Technology, China\\
   \texttt{Tengfei.Liu0821@outlook.com} \\
   \And
   Mingjie Li \\
   Stanford University, USA\\
   \texttt{lmj695@gmail.com} \\
   \And
   Hanyu Gu\\
   University of Technology Sydney, Australia\\
  \texttt{hanyu.gu@uts.edu.au}\\
     \And
   Hui Zhang \\
   Hunan University, China\\
   \texttt{zhanghuihby@126.com} \\
  \And
   Xiaojun Chang \\
   University of Technology Sydney, Australia\\
   \texttt{Xiaojun.Chang@uts.edu.au} \\
   \And
  Yaonan Wang \\
   Hunan University, China\\
   \texttt{yaonan@hnu.edu}
}

\begin{document}

\maketitle

\begin{abstract}
Knowledge distillation (KD) is a widely adopted and effective method for compressing models in object detection tasks. Particularly, feature-based distillation methods have shown remarkable performance. Existing approaches often ignore the uncertainty in the teacher model's knowledge, which stems from data noise and imperfect training. This limits the student model's ability to learn latent knowledge, as it may overly rely on the teacher's imperfect guidance. In this paper, we propose a novel feature-based distillation paradigm with knowledge uncertainty for object detection, termed "\textbf{U}ncertainty Estimation-Discriminative Knowledge \textbf{E}xtraction-Knowledge \textbf{T}ransfer (\textbf{UET})", which can seamlessly integrate with existing distillation methods. By leveraging the Monte Carlo dropout technique, we introduce knowledge uncertainty into the training process of the student model, facilitating deeper exploration of latent knowledge. Our method performs effectively during the KD process without requiring intricate structures or extensive computational resources. Extensive experiments validate the effectiveness of our proposed approach across various distillation strategies, detectors, and backbone architectures. Specifically, following our proposed paradigm, the existing FGD  method achieves state-of-the-art (SoTA) performance, with ResNet50-based GFL achieving 44.1\% mAP on the COCO dataset, surpassing the baselines by 3.9\%.
\end{abstract}

\section{Introduction}
Deep learning (DL) has demonstrated remarkable dominion in various fundamental visual tasks \cite{1}, such as object detection \cite{2}. However, the scale and parameterization of models restrict the practical application of DL-based detectors in real-world tasks. To address this, several techniques have been proposed, including pruning \cite{3}, weight quantization \cite{4}, and knowledge distillation \cite{5}. Knowledge distillation (KD), as a means of model compression, aims to utilize well-trained teacher models to guide student model learning for knowledge transfer.

\begin{figure}
  \centering
  \includegraphics[scale=0.3]{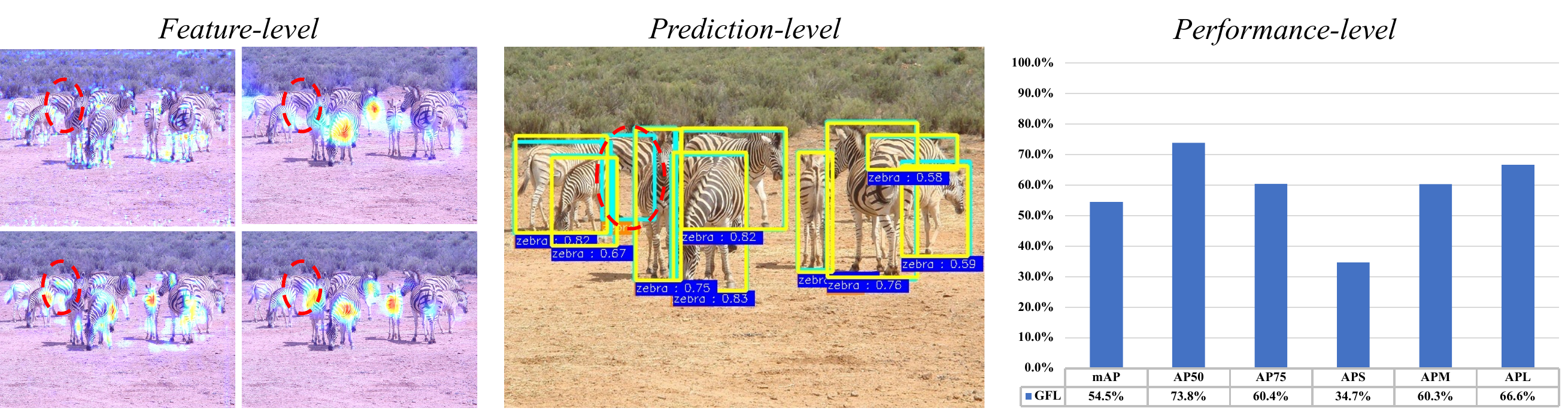}
  \caption{We chose the popular ResNet-101 GFL framework which trained with 24 epochs and employed multi-scale training on the COCO dataset as the teacher detector. Subsequently, we assess its performance on the training set of the COCO dataset. The heatmap on the left showcases the multi-scale features of the GFL detector, while the middle segment illustrates the prediction results of GFL. On the right, we present the evaluation performance of the GFL.}
  \label{intro}
\end{figure}
In object detection tasks, aligning the intermediate representations (feature-level) of the student and teacher models to achieve knowledge transfer is a common practice \cite{6,7,8, 9}. In feature-based KD, a paradigm of “Discriminative Knowledge \textbf{E}xtraction-Knowledge \textbf{T}ransfer (\textbf{ET})” is typically followed. Object-related knowledge extraction involves designing specific modules to extract effective knowledge from the intermediate features of the teacher model, such as structural knowledge \cite{8}, and instance-level fine-grained knowledge \cite{9}. Knowledge transfer involves a loss function that constrains the distance between the extracted knowledge of the student and teacher models, like the $l_p$-norm distance. For a specific example, FGD \cite{7} uses a GC-block module \cite{10}  to extract the global relationship knowledge of pixel-to-pixel from the intermediate features of the teacher model and then transfers this knowledge to the student model. 

 For teacher detectors, the knowledge they acquire inherently exhibits uncertainty. This arises primarily because the teacher detector's knowledge is derived from learning through training data and parameter optimization. Despite meticulous annotation, the training data for detection can still introduce noise due to subjective judgments by labeling staff, which is referred to as aleatoric uncertainty \cite{28}, especially in complex scenarios such as occlusions. Moreover, the design of the detector and the inherent randomness in the training process contribute to epistemic uncertainty \cite{28}, further complicating the knowledge acquired. Essentially, the knowledge possessed by teacher detectors stems from their previous learning experiences, which are inherently imperfect. Therefore, epistemic and aleatoric uncertainties collectively contribute to the knowledge uncertainties.

In reality, uncertainty is an intrinsic characteristic of knowledge, significantly impacting the process of knowledge transfer \cite{46}. To further explore the uncertainty of teachers' knowledge, we evaluate the popular teacher detector (GFL \cite{14}) on the train set of the MS COCO \cite{12}. Surprisingly, the mean Average Precision (mAP) of the teacher detector only reached 54.4\%, as shown in Figure \ref{intro} (right). We notice inconsistencies in the model's multi-scale feature heatmaps in complex scenarios, such as occlusions. For instance, in the first row of the heatmap (Figure \ref{intro}, left), the teacher detector fails to focus on the red-highlighted area, whereas in the second row, it does. This inconsistency of knowledge may confuse the teacher detector, potentially leading to missed detections (Figure \ref{intro} (middle)).
Ignoring knowledge uncertainty and directly proceeding with knowledge extraction and transfer can be unwise, as it may cause the student model to be misled by the teacher's unreliable knowledge. Explicitly quantifying uncertainty has shown benefits in various areas, such as segmentation \cite{37} and localization \cite{47}. 
Existing approaches typically focus on using elegant formulas to quantify uncertainty. Consequently, the success of these methods is highly dependent on the accuracy of the quantification techniques. However, representing quantified uncertainty during the KD process poses challenges, and these imperfect approximation methods may restrict the student's learning potential.

%\textbf{Therefore, implicitly incorporating uncertainty offers a promising alternative that aligns more effectively with the knowledge transfer process.}
In this work, we argue that considering the uncertainty of the teacher's knowledge seriously is crucial in the process of KD for object detection. To address this, we propose a novel paradigm with knowledge uncertainty for feature-based distillation detection, termed \textbf{U}ncertainty Estimation-Discriminative Knowledge \textbf{E}xtraction-Knowledge \textbf{T}ransfer (\textbf{UET}). Unlike previous approaches, we model the KD process with knowledge uncertainty. And, we propose an uncertainty knowledge estimate method for the KD process. We estimate the uncertainty knowledge of the teacher models using Monte Carlo dropout (MC dropout) \cite{11} before knowledge extraction, and then incorporate this knowledge into the training process of the student model. This strategy mitigates the misleading risk of the student model caused by the overconfidence of the teacher model, thereby enhancing the training effectiveness of the student model. Notably, our proposed paradigm can be seamlessly integrated with other distillation methods, offering substantial improvements with minimal computational overhead. We conduct extensive experiments on the MS COCO to verify the effectiveness of our proposed method across various distillation strategies, detectors, and backbone architectures. When adopting FGD as the baseline in our paradigm, ResNet50-based \cite{13} detectors GFL \cite{14}, Faster R-CNN \cite{17}, RetinaNet \cite{15}, and FCOS \cite{16} can achieve 44.1\% mAP, 40.8\% mAP, 39.9\% mAP, and 42.9\% mAP respectively on the COCO dataset, which surpasses the previous advanced KD methods. 
%Last but not least, most work focuses on quantifying the uncertainty, we find that even if we only introduce the knowledge uncertainty can also considerably profit from the during the training stage. 
In summary, the contributions of this paper can be summarized:
\begin{itemize}
    \item We propose a new paradigm of "\textbf{U}ncertainty Estimation-Discriminative Knowledge \textbf{E}xtraction-Knowledge \textbf{T}ransfer", for the feature-based distillation methods. This approach introduces knowledge uncertainty into the learning process of the student model, reducing the risk of being misled by the teacher model. \textbf{UET} seamlessly integrates with existing distillation methods, delivering improvements without any additional complexity.
\end{itemize}
\begin{itemize}
    \item We propose a straightforward yet effective method for uncertainty estimation by combining Monte Carlo dropout, enabling the capture of the teacher model's uncertainty during the knowledge distillation process.
\end{itemize}
\begin{itemize}
    \item We conduct extensive experiments on the COCO dataset to verify the effectiveness of the proposed paradigm across various distillation strategies, detectors, and backbone architectures, achieving SoTA performance. 
\end{itemize}

\section{Related works}
\subsection{Object detection with knowledge distillation}
Object detection is one of the most fundamental tasks in image processing. Due to the advancements in deep learning, detectors based on convolutional neural networks (CNNs) have achieved remarkable results  \cite{14,15,16,17}. Currently, CNN-based detectors can be categorized into two-stage detectors and one-stage detectors, depending on whether they are equipped with a Region Proposal Network (RPN). Two-stage detectors, such as Faster R-CNN \cite{17}, utilize RPN for foreground and background object identification, enabling finer detection. On the other hand, one-stage models directly predict targets densely using multi-scale features extracted by FPN \cite{18}, as seen in RetinaNet \cite{15} and GFL \cite{14}. Additionally, recent developments have introduced anchor-free methods (like FCOS \cite{16}) to reduce the detector's reliance on anchors. In this work, we explore the effectiveness of the proposed method across different types of detectors.

Knowledge distillation aims to transfer knowledge from a well-trained but cumbersome teacher detector to a lightweight student detector. Currently, KD in object detection can be categorized into feature-based distillation \cite{6,7,8,9,22,23, 24, 25} and logit-based distillation \cite{19, 20, 21, 26}. Due to its simpler and more uniform form, feature-based knowledge distillation is popular in object detection. Presently, feature-based knowledge distillation follows the paradigm of "\textbf{ET}". Most works focus on the design of discriminative knowledge extraction modules (such as scale-aware knowledge \cite{24}, relationship-related knowledge \cite{16}, etc.) and transfer methods (Pearson Correlation Coefficient \cite{6}, SSIM \cite{25}, etc.). In contrast to other works, we consider the uncertainty of knowledge in the teacher model and propose a novel \textbf{UET} paradigm for feature-based distillation. We aim to improve the existing feature distillation procedure by following \textbf{UET} to help the student model explore potential knowledge.

\subsection{Uncertainty modeling}
Uncertainty often arises from insufficient knowledge and data during model training, prompting the need for robust uncertainty estimation methods to quantify prediction reliability \cite{11, 27, 28, 40, 41, 42, 43, 44, 45}.  Recent research has witnessed a surge in exploring uncertainty in the knowledge distillation task. For instance, the UNIX \cite{36} proposed to reduce computation costs by combining uncertainty sampling and adaptive mixup to prioritize informative samples.  Uncertainty Distillation method \cite{37}, on the other hand, aimed to quantify prediction uncertainty by training a compact model to mimic the output distribution of a large ensemble of models, enabling efficient and reliable uncertainty estimation. Similarly, the Uncertainty-aware Contrastive Distillation (UCD) \cite{38} method attempts to alleviate catastrophic forgetting in incremental semantic segmentation by contrasting features between new and frozen models. Avatar Knowledge Distillation (AKD) \cite{39} strived to generate inference ensemble models (Avatars) from a teacher model and adaptively adjust their contribution to knowledge transfer using an uncertainty-aware factor, refining the previous distillation methods for dense prediction tasks. Despite these advances, existing uncertainty estimation methods often require fine-grained modeling, and there has been limited exploration in the object detection domain. To address this limitation, we introduce knowledge uncertainty into the KD framework for object detection, guiding students to learn more potential knowledge to enhance its detection performance. 

% Bayesian Neural Networks (BNNs) \cite{40} are notable for their ability to interpret posterior uncertainties of model parameters. This is achieved through techniques like weight sampling \cite{42, 43} and multiple forward passes \cite{11, 44, 45} through the network, enabling the assessment of prediction uncertainties.

% By leveraging uncertainty modeling techniques across these diverse tasks, researchers aim to improve the robustness and interpretability of deep learning models, while improving their performance. However, existing uncertainty estimation methods often require fine-grained modeling, and there has been limited exploration in the object detection domain. Therefore, in this paper, we further introduce uncertainty knowledge into the knowledge distillation framework for object detection tasks, guiding the student network to learn more robust knowledge to enhance its detection performance.

\section{Methods}
\begin{figure}
  \centering
  \includegraphics[scale=0.28]{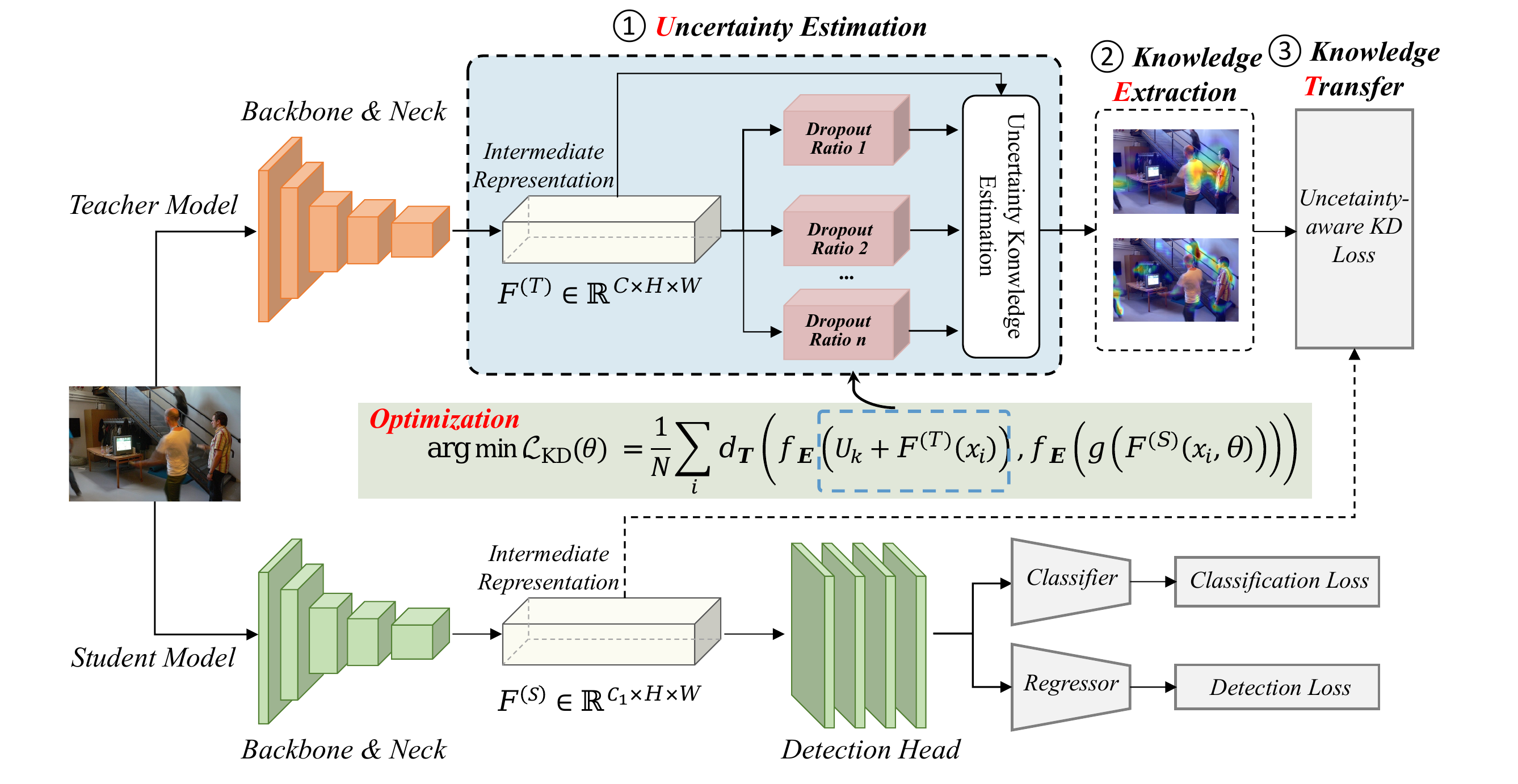}
  \caption{The overview of the proposed \textbf{UET} paradigm.}
  \label{method}
\end{figure}

In this section, we first present the conventional \textbf{ET} paradigm of feature distillation. Next, we formulate our proposed \textbf{UET} paradigm for feature distillation, which aims to alleviate the problem of misleading due to excessive reliance on teacher knowledge. Additionally, we provide the pseudocode of the \textbf{UET} paradigm in the Appendix \ref{A3}. Finally, we introduce an uncertainty estimation method based on MC dropout, which allows for the simple yet effective incorporation of knowledge uncertainty during the KD process. The overview of \textbf{UET} is represented in Figure \ref{method}.

\subsection{KD model in \textbf{ET} paradigm}
In object detection, we typically construct a multi-scale feature using the FPN network \cite{18} to enhance the detector's perception of features at different scales. In feature-based knowledge distillation, knowledge transfer usually occurs on the multi-scale features $F=[F_1, \ldots, F_M]$, where $F_i$ is the $i$-th  scale feature after passing an image through the FPN network. The typical value of $M$ in the literature is 5. Conventional feature-based distillation methods follow the \textbf{ET} paradigm. Specifically, knowledge extraction methods $f_\textbf{E}(\cdot)$ are first used to extract the discriminative knowledge that the student model is expected to learn from the multi-scale features of the teacher model, such as the GC module \cite{10} used in the FGD method \cite{7} for extracting pixel-wise relationships. Then, a designed constraint function $d_\textbf{T}(\cdot)$ is used to minimize the difference between the discriminative knowledge of the teacher and student models, such as the SSIM \cite{25}, to complete knowledge transfer. This process can be formalized as:

% \begin{equation}
% K^{(T)}_L(x)= f_\textbf{E}(F^{(T)}_L(x)), K^{(S)}_L(x)=f_\textbf{E}(g(F^{(S)}_L(x, \theta))),
% \end{equation}

% \begin{equation}
% \arg\min_{\theta}  \mathcal{L}_{\text{KD}} (\theta) = d_\textbf{T}(K^{(T)}_L(x), K^{(S)}_L(x, \theta)),
% \end{equation}

\begin{equation}
\arg\min_{\theta}  \mathcal{L}_{\text{KD}} (\theta) =  \frac{1}{N}\sum_i d_\textbf{T}(f_\textbf{E}(F^{(T)}(x_i)), f_\textbf{E}(g(F^{(S)}(x_i, \theta)))),
\end{equation}

where $x_i$ is the input image from training data, $\theta$ is the parameters of the student detector, $F^{(T)}(x)$ and $F^{(S)}(x)$ represent the FPN network's output features of the teacher and student models, respectively. Additionally, $g(x)$ denotes the adaptive function typically used to align the multi-scale features of the teacher and student models. It is worth mentioning that during the training process, only the parameters of the student model are updated through back-propagation.

\subsection{Formulation of KD with knowledge uncertainty}
As mentioned earlier, the inevitable uncertainty in the knowledge of the teacher model arises from data noise and the randomness of training. However, the \textbf{ET} paradigm overlooks the inherent uncertainty in the knowledge of the teacher model, which may cause the student model to be misled by the overconfidence in the teacher's knowledge.
Assuming that the knowledge of the teacher model, $(\textbf{x},\textbf{F}^{(T)})$, follows a distribution $\mathcal{U}$ on $\mathcal{X}\times \mathcal{Y}$, where $\mathcal{X}$ is the instance space of input images and $\mathcal{Y}$ is the knowledge space. Based on this, the KD process of \textbf{UET} paradigm can be modeled as follows:

\begin{equation}
\arg\min_{\theta}  \mathcal{L}_{\text{KD}} (\theta) = \mathbb{E}_\mathcal{U}\{d_\textbf{T}(f_\textbf{E}(\textbf{F}^{(T)}), f_\textbf{E}(g(F^{(S)}(\textbf{x}, \theta))))\}.
\label{2}
\end{equation}
However, due to the intractability of computing the expectation for high dimensional distribution, we consider the approximation

\begin{equation}
\arg\min_{\theta}  \overline{\mathcal{L}}_{\text{KD}} (\theta) = \mathbb{E}_{\mathcal{U}_x}\{d_\textbf{T}(f_\textbf{E}(\mathbb{E}_{\mathcal{U}_{|x}}\{\textbf{F}^{(T)}\}, f_\textbf{E}(g(F^{(S)}(\textbf{x}, \theta))))\},
\end{equation}
where $\mathcal{U}_x$ is the marginal distribution of $x$, and $\mathcal{U}_{|x}$ is the conditional distribution of the knowledge on $x$. Since
 $\mathcal{U}_{|x}$ is unknown in practice, it can be approximated if a sample $\{F^{(T)}_{i}(x)|i = 1, \ldots, N\}$ is available. Then, we have
 
\begin{equation}
\arg\min_{\theta}  \widehat{\mathcal{L}}_{\text{KD}} (\theta) = \mathbb{E}_{\mathcal{U}_x}\{d_\textbf{T}(f_\textbf{E}(\frac{1}{N}\sum_i^N F^{(T)}_{i}(x)), f_\textbf{E}(g(F^{(S)}(\textbf{x}, \theta))))\}.
\label{4}
\end{equation}

From Equation \ref{4}, we formulate the KD process with knowledge uncertainty. By minimizing the objective function in Equation \ref{4}, we can incorporate the uncertainty of teacher model knowledge into the optimization process of the student model.

\subsection{Uncertainty estimation}
After formulating the feature distillation with knowledge uncertainty, we can make use of the teacher's knowledge uncertainty in the training process of the student model. 
However, the conditional distribution of the teacher's knowledge is unknown, which results in sampling from this distribution is also challenging. Modeling the uncertainty of the deep learning model has been a topic of considerable attention \cite{27, 28}. Especially, \cite{11} establishes a theoretical connection between dropout training in deep neural networks and Bayesian inference in deep Gaussian processes. More importantly, they introduced a Monte Carlo estimate method based on the dropout, referred to as MC dropout, which can realize the estimating of uncertainty. There has been a lot of work \cite{39, 48, 49} estimating uncertainty through MC dropout and achieving outstanding performance in specific fields. 
Inspired by this, we utilize MC dropout to estimate the uncertain knowledge of the teacher models. More specifically, we perform $N$-times the teacher's forward process with dropout and denote the teacher's knowledge of the $i$-th run as $Dropout_i^{(T)}(x)$. The set $\{Dropout_i^{(T)}(x)|i = 1, \ldots, N\}$ is used as a sample of the uncertainty knowledge $\textbf{F}^{(T)}$ as in Equation \ref{4}. Then, we have
\begin{equation}
U_K = \frac{1}{N}\sum_i^N F^{(T)}_{i}(x) \approx \frac{1}{N} \sum_{i=1}^{N} Dropout_i^{(T)}(x).
\label{5}
\end{equation}

It is impractical and resource-intensive in real-world scenarios with a large sample size of $N$. 
Due to the limitations in the number of samples during the actual implementation process, the estimated uncertainty knowledge $U_k$ may have a large variance which could potentially limit the effectiveness of our proposed method in knowledge transfer. To mitigate this, we combine the original teacher knowledge with the estimated uncertainty knowledge $U_K$ (similar to residual learning), enhancing the student's training process.  Our proposed \textbf{UET} paradigm for feature distillation in object detection can be summarized as:
\begin{equation}
\arg\min_{\theta}  \mathcal{L}_{\text{KD}} (\theta) = \frac{1}{N}\sum_i d_\textbf{T}(f_\textbf{E}(U_K + F^{(T)}(x_i)), f_\textbf{E}(g(F^{(S)}(x_i, \theta)))).
\label{6}
\end{equation}

Compared to the traditional \textbf{ET} paradigm, this approach incorporates knowledge uncertainty into the student's learning process in a straightforward manner. Unlike the complex uncertainty quantification methods typically used, our approach does not require explicit quantification of knowledge uncertainty. Instead, we introduce a simple yet effective method that integrates knowledge uncertainty into the student model's training with minimal computational cost, which is almost negligible. Additionally, our method functions as a plug-and-play solution for other distillation models, enhancing the performance of the student model without additional complexity.

\section{Experiments}
\label{4.0}

\begin{table}[!t]
  \centering
  \caption{Results in different type framework on MS COCO.}
    \begin{tabular}{lccccccc}
    \toprule
    \multicolumn{1}{c}{Method} & Schedule & mAP   & $AP_{50}$  & $AP_{75}$ & $AP_S$  & $AP_M$   & $AP_L$ \\
    \midrule
    \multicolumn{8}{c}{\textit{Two-stage detectors}} \\
    \rowcolor[rgb]{ .906,  .902,  .902} Faster R-CNN-Res101 (T) & 2$\times$    & 39.8  & 60.1  & 43.3  & 22.5  & 43.6  & 52.8 \\
    \rowcolor[rgb]{ .906,  .902,  .902} Faster R-CNN-Res50 (S) & 2$\times$    & 38.4  & 59.0  & 42.0  & 21.5  & 42.1  & 50.3 \\
    FitNet \cite{30} & 2$\times$    & 38.9  & 59.5  & 42.4  & 21.9  & 42.2  & 51.6 \\
    FRS \cite{34}  & 2$\times$    & 39.5  & 60.1  & 43.3  & 22.3  & 43.6  & 51.7 \\
    FGD  \cite{7} & 2$\times$    & 40.5  & -     & -     & 22.6  & 44.7  & 53.2 \\
    DiffKD \cite{22} & 2$\times$    & 40.6  & 60.9  & 43.9  & 23.0  & 44.5  & \textbf{54.0} \\    
    Ours+FGD & 2$\times$    & \textbf{40.8 (+2.4)}  & \textbf{61.0 }& \textbf{44.5}  & \textbf{23.5}  & \textbf{44.9 } & 53.7  \\
    \midrule
    \multicolumn{8}{c}{\textit{One-stage detectors}} \\
    \rowcolor[rgb]{ .906,  .902,  .902} RetinaNet-Res101 (T) & 2$\times$    & 38.9  & 58.0  & 41.5  & 21.0  & 42.8  & 52.4 \\
    \rowcolor[rgb]{ .906,  .902,  .902} RetinaNet-Res50 (S) & 2$\times$   & 37.4  & 56.7  & 39.6  & 20.6  & 40.7  & 49.7 \\
    FitNet \cite{30}& 2$\times$    & 37.4  & 57.1  & 40.0  & 20.8  & 40.8  & 50.9 \\
    FRS \cite{34}  & 2$\times$    & 39.3  & 58.8  & 42.0  & 21.5  & 43.3  & 52.6 \\
    CrossKD \cite{20}& 2$\times$    & 39.7  & 58.9  & 42.5  & \textbf{22.4}  & 43.6  & 52.8 \\
    FGD  \cite{7} & 2$\times$    & 39.7  & -     & -     & 22.0  & 43.7  & \textbf{53.6} \\
    DiffKD \cite{22}& 2$\times$    & 39.7  & 58.6  & 42.1  & 21.6  & 43.8  & 53.3 \\
    Ours+FGD & 2$\times$    & \textbf{39.9 (+2.5)} & \textbf{59.0} & \textbf{42.7} & 22.1 & \textbf{43.9} & 53.4 \\
    \midrule
    \multicolumn{8}{c}{\textit{Anchor-free detectors}} \\
    \rowcolor[rgb]{ .906,  .902,  .902} FCOS-Res101 (T) & 2$\times$, ms    & 40.8  & 60.0  & 44.0  & 24.2  & 44.3  & 52.4 \\
    \rowcolor[rgb]{ .906,  .902,  .902} FCOS-Res50 (S) & 2$\times$, ms    & 38.5  & 57.7  & 41.0  & 21.9  & 42.8  & 48.6 \\
    FRS  \cite{34} & 2$\times$    & 40.9  & 60.3  & 43.6  & 25.7  & 45.2  & 51.2 \\
    CrossKD \cite{20}& 2$\times$   & 41.3  & 60.6  & 44.2  & 25.1  & 45.5  & 52.4 \\
    DiffKD \cite{22}& 2$\times$    & 42.4  & 61.0  & 45.8  & 26.6  & 45.9  & 54.8 \\
    FGD  \cite{7} & 2$\times$   & 42.7  & -     & -     & \textbf{27.2} & 46.5  & \textbf{55.5} \\
    Ours+FGD & 2$\times$    & \textbf{42.9 (+4.4)} & \textbf{61.6} & \textbf{46.3} & 27.1  & \textbf{46.8} & 54.7 \\
    \bottomrule
    \end{tabular}%
  \label{tab2}%
\end{table}%

\subsection{Settings}
\label{4.1}
To validate the effectiveness of our methods, we conduct extensive experiments on the MS-COCO dataset across various detectors, KDs, and different backbones. We use the 120k images of datasets to train the models and 5k val images for evaluating the experiments. We report mean Average Precision (mAP) as the evaluation metric along with AP at different scales and thresholds, including $AP_{50}$, $AP_{75}$, $AP_S$, $AP_M$, and $AP_L$. 
All experiments are conducted on a machine equipped with 2 NVIDIA GeForce RTX 3090 GPUs. Additionally, we select FGD \cite{7} as our baseline, completing the "\textbf{ET}" process in our proposed "\textbf{UET}" paradigm. All experiments were implemented using the mmdetection \cite{29} and PyTorch framework. Except for setting the batch size to 8, we follow the training settings of FGD \cite{7}. In this paper, 1$\times$, 2$\times$, and ms respectively denote 12 epochs, 24 epochs, and multi-scale training. Unless otherwise specified, we set $N$=5 in our experiments. The first group employs a ratio strategy with an initial ratio of 0.05 and a common difference of 0.05. In these experiments, we introduce knowledge uncertainty solely on the teacher's side.

\subsection{Main results}
\label{4.2}

\textbf{Comparison experiments with other detectors.}
To explore the possibility of our proposed method across different types of detectors, we conduct comparative experiments on Faster R-CNN (two-stage detector), RetinaNet (anchor-based detector), and FCOS (anchor-free detector). The experimental results, as presented in Table \ref{tab2}, demonstrate that our method, when incorporated on top of FGD, achieves SoTA performance across three different types of detectors. Specifically, within the Faster R-CNN framework, we attain a 40.8\% mAP, marking a 2.4\% improvement over the student model. Similarly, in the RetinaNet detector, we achieve a 39.9\% mAP, representing a 2.5\% improvement over the student model. Furthermore, within the FCOS detector, we achieve a 42.9\% mAP, reflecting a 4.4\% improvement over the student model. 

\textbf{Comparison experiments with GFL framework.} To further demonstrate the superiority of our approach, we conduct comparisons with the previous SoTA KD methods on the popular GFL framework. In this experiment, we utilize a ResNet101 backbone for the GFL, trained for 24 epochs with multi-scale training, as our teacher network. As for the student model, we employ a ResNet50 backbone for the GFL, trained for 12 epochs. Without resorting to any elaborate techniques, our method achieves new SoTA performance, as depicted in Table \ref{tab1}. Specifically, our approach achieves an mAP of 44.1\%, surpassing the previous SoTA distillation methods CrossKD \cite{20} (43.7\% mAP) and BCKD \cite{21} 
 (43.2\% mAP), as well as the baseline FGD \cite{7} (43.4\% mAP). Moreover, our method exhibited a notable improvement of 3.9\% over the original student model (40.2\% mAP).

These experiments validate that following our proposed \textbf{UET} paradigm effectively enhances the learning potential of student detectors during the KD process. Furthermore, we demonstrate that our paradigm is adaptable to various styles of detectors.

\begin{table}[!t]
    \centering
    \begin{minipage}[t]{0.49\textwidth}
        \centering
        \caption{Comparison results in GFL framework on MS COCO. * denotes our reproduced results. We set $N$ to 10, choose the  \textbf{\textit{b)}} as ratio strategy, and only introduce knowledge uncertainty into the teacher side in this experiment.}
        \setlength{\tabcolsep}{0.5mm}
\renewcommand{\arraystretch}{1.1}
\resizebox{1.0\textwidth}{!}{
        \begin{tabular}{lcccccc}
    \toprule
    \multicolumn{1}{c}{Method} & mAP   & $AP_{50}$  & $AP_{75}$ & $AP_S$  & $AP_M$   & $AP_L$ \\
    \midrule
    \rowcolor[rgb]{ .906,  .902,  .902} GFL-Res101 (T) & 44.9  & 63.1  & 49.0  & 28.0  & 49.1  & 57.2 \\
    \rowcolor[rgb]{ .906,  .902,  .902} GFL-Res50 (S) & 40.2  & 58.4  & 43.3  & 23.3  & 44.0  & 52.2 \\
    \midrule
    FitNets \cite{30} & 40.7  & 58.6  & 44.0  & 23.7  & 44.4  & 53.2 \\
    Inside GT Box & 40.7  & 58.6  & 44.2  & 23.1  & 44.5  & 53.5 \\
    Defeat \cite{31}& 40.8  & 58.6  & 44.2  & 24.3  & 44.6  & 53.7 \\
    %LD \cite{19}   & 41.0  & 58.6  & 44.2  & 23.4  & 45.0  & 53.1 \\
    Main Region \cite{19} & 41.1  & 58.7  & 44.4  & 24.1  & 44.6  & 53.6 \\
    Fine-Grained \cite{32}& 41.1  & 58.8  & 44.8  & 23.3  & 45.4  & 53.1 \\
    GID \cite{33}  & 41.5  & 59.6  & 45.2  & 24.3  & 45.7  & 53.6 \\
    SKD \cite{25}  & 42.3  & 60.2  & 45.9  & 24.4  & 46.7  & 55.6 \\
    ScaleKD \cite{24}& 42.5  & -     & -     & 25.9  & 46.2  & 54.6 \\
    BCKD \cite{21} & 43.2  & 61.6  & 46.9  & 25.7  & 47.3  & 55.9 \\
    FGD* \cite{7}  & 43.4  & 61.7  & 47.0  & 26.2  & 47.4  & 56.4 \\
    CrossKD \cite{20}& 43.7  & 62.1  & 47.4  & \textbf{26.9} & 48.0  & 56.2 \\
    %Ours  & \textbf{44.0} & \textbf{62.2} & \textbf{47.7} & 26.8  & \textbf{48.4} & \textbf{57.0} \\
    \midrule
    Ours+FGD  & \textbf{44.1} & \textbf{62.3} & \textbf{47.8} & 26.6  & \textbf{48.2} & \textbf{56.9} \\
    \bottomrule
    \end{tabular}}
    \label{tab1}
    \end{minipage}
    \hfill
    \begin{minipage}[t]{0.49\textwidth}
        \centering
        \caption{Sensitivity analysis of the ratios strategy. The $N$ is uniformly set to 5 in this experiment.}
        \setlength{\tabcolsep}{0.2mm}
\renewcommand{\arraystretch}{1.4}
        \begin{tabular}{ccccccc}
    \toprule
    Strategy & mAP  & $AP_{50}$  & $AP_{75}$ & $AP_S$  & $AP_M$   & $AP_L$ \\
    \midrule
    \textbf{\textit{a)}} & \textbf{44.0}  & 62.2  & 47.8  & 27.0  & 48.3  & 56.7 \\
    \textbf{\textit{b)}}     & \textbf{44.0}  & 62.3  & 47.5  & 26.7  & 48.2  & 57.0 \\
    \textbf{\textit{c)}}   & \textbf{44.0}  & 62.4  & 47.8  & 26.9  & 48.2  & 56.9 \\
    \bottomrule
    \end{tabular}%
        \label{tab3}
    \vspace{0.4cm}
    \centering
  \caption{Sensitivity study of the $N$. We employ the strategy \textbf{\textit{b)}} as the ratio policy.}
  \setlength{\tabcolsep}{0.8mm}
\renewcommand{\arraystretch}{1.3}
    \begin{tabular}{ccccccc}
    \toprule
    N     & mAP   & $AP_{50}$  & $AP_{75}$ & $AP_S$  & $AP_M$   & $AP_L$ \\
    \midrule
    0     & 43.4  & 61.7  & 47.0  & 26.2  & 47.4  & 56.4 \\
    1     & 43.9  & 62.1  & 47.5  & 26.4  & 48.1  & 57.0 \\
    5     & 44.0  & 62.3  & 47.5  & 26.7  & 48.2  & 57.0 \\
    10    & \textbf{44.1}  & 62.3  & 47.8  & 26.6  & 48.2  & 56.9 \\
    15    & 43.8  & 61.9  & 47.6  & 26.5  & 48.0  & 57.0 \\
    \bottomrule
    \end{tabular}%
  \label{tab4}%
    \end{minipage}
\end{table}

\subsection{Ablation analysis}
In our proposed method for estimating knowledge uncertainty, dropout technology plays a crucial role. Therefore, in this part, we primarily investigate the impact of dropout on student learning, such as dropout ratio and frequency. Here we adopt the GFL framework, and the experimental settings are consistent with Section \ref{4.2}. Next, we will delve into these aspects in detail.

\textbf{Sensitivity analysis of the ratios strategy.} We design three different strategies to investigate the impact of dropout ratio on the model: \textbf{\textit{a)}} where $N$ groups of ratios are all fixed at 0.15; \textbf{\textit{b)}}  where the first group has a ratio of 0.05 with a common difference of 0.05; \textbf{\textit{c)}} which builds upon \textbf{\textit{b)}} by increasing with epochs, at a growth rate of 0.025 per epoch. These results, as shown in Table \ref{tab3}, indicate that our network is not sensitive to different ratio strategies. Particularly, when adopting strategy \textbf{\textit{c)}}, after 10 epochs of training, with a dropout ratio ranging from 0.3 to 0.5, our method still maintains an mAP of 44.0\%.

% % Table generated by Excel2LaTeX from sheet 'Sheet1'
% \begin{table}[htbp]
%   \centering
%   \caption{Sensitivity analysis of the ratios strategy.}
%     \begin{tabular}{ccccccc}
%     \toprule
%     Ratio Strategy & mAP   & $AP_{50}$  & $AP_{75}$ & $AP_S$  & $AP_M$   & $AP_L$ \\
%     \midrule
%     \textbf{\textit{a)}} & 44.0  & 62.2  & 47.8  & 27.0  & 48.3  & 56.7 \\
%     \textbf{\textit{b)}}     & 44.0  & 62.3  & 47.5  & 26.7  & 48.2  & 57.0 \\
%     \textbf{\textit{c)}}   & 44.0  & 62.4  & 47.8  & 26.9  & 48.2  & 56.9 \\
%     \bottomrule
%     \end{tabular}%
%   \label{tab3}%
% \end{table}%

\textbf{Sensitivity study of the $N$.} To explore the influence of the sampling count $N$ in MC dropout on the model, we conduct comparative experiments using different $N$ values. The experimental results, as presented in Table \ref{tab4}, demonstrate the beneficial effect of introducing uncertainty from the teacher model to the student model. Even with just one dropout, the model's performance reaches 43.9\%, a 0.5\% improvement over the pure FGD method. When increasing $N$ to 10, we observe the model's mAP reaching 44.1\%, a 0.7\% enhancement compared to the pure FGD method. However, with a further increase in $N$ to 15, the model's mAP reaches 43.8\%, a slight decrease of 0.3\% compared to $N=10$, yet still outperforming the FGD method by 0.4\%.

% % Table generated by Excel2LaTeX from sheet 'Sheet1'
% \begin{table}[htbp]
%   \centering
%   \caption{Sensitivity study of the times.}
%     \begin{tabular}{ccccccc}
%     \toprule
%     N     & mAP   & $AP_{50}$  & $AP_{75}$ & $AP_S$  & $AP_M$   & $AP_L$ \\
%     \midrule
%     0     & 43.4  & 61.7  & 47.0  & 26.2  & 47.4  & 56.4 \\
%     1     & 43.9  & 62.1  & 47.5  & 26.4  & 48.1  & 57 \\
%     5     & 44.0  & 62.3  & 47.5  & 26.7  & 48.2  & 57.0 \\
%     10    & 44.1  & 62.3  & 47.8  & 26.6  & 48.2  & 56.9 \\
%     15    & 43.8  & 61.9  & 47.6  & 26.5  & 48.0  & 57.0 \\
%     \bottomrule
%     \end{tabular}%
%   \label{tab4}%
% \end{table}%

\textbf{Effect of knowledge sources.}
We conduct comparative experiments on introducing uncertainty from different knowledge sources. As shown in Table \ref{tab5}, when uncertainty is exclusively integrated into the teacher's knowledge, the mAP reaches 44.0\%. Similarly, when it is introduced solely into the student knowledge, the mAP is 43.6\%. In both cases, these values represent enhancements of 0.6\% and 0.2\% respectively compared to the pure FGD method. This suggests that incorporating knowledge uncertainty into the KD process benefits the learning process of the student model. However, when knowledge uncertainty is simultaneously introduced to the teacher and student, the model's mAP remains at 44.0\%. In this case, simultaneously handling both types of uncertain knowledge may have increased learning complexity, making the optimization process more challenging.

\textbf{Effect of Residual structure.} We also explore the impact of residual structures, and the results are shown in Table \ref{tab5}. When we do not introduce the residual structure, the student detector achieves an mAP of 43.5\%, which is 0.5\% lower than after introducing it. In the presence of a constrained sample size, our methodology may not fully manifest its efficacy. This also indicates that incorporating knowledge uncertainty estimation with residual structures is more beneficial to the learning process of the student detector.

\begin{table}[!t]
  \centering
  \caption{Effect of knowledge sources. Teacher: the uncertainty is exclusively integrated into the teacher's knowledge, Student: the uncertainty is exclusively integrated into the student's knowledge, Residual: introduce the residual learning.}
    \begin{tabular}{ccccccccc}
    \toprule
    Teacher & Student & Residual & mAP   & $AP_{50}$  & $AP_{75}$ & $AP_S$  & $AP_M$   & $AP_L$ \\
    \midrule
          &      &  & 43.4  & 61.7  & 47.0  & 26.2  & 47.4  & 56.4 \\
          &$\checkmark$  & $\checkmark$  & 43.6  & 61.8  & 47.3  & 26.1  & 47.8  & 56.8 \\
     $\checkmark$      &  &   $\checkmark$  & \textbf{44.0}  & 62.2  & \textbf{47.7}  & \textbf{26.8}  & \textbf{48.4}  & \textbf{57.0} \\
     $\checkmark$      &   &    & 43.5  & 61.7  & 46.9  & 26.0  & 47.7  & 56.4 \\
     $\checkmark$    & $\checkmark$  & $\checkmark$  & \textbf{44.0}  & \textbf{62.3}  & 47.5  & 26.7  & 48.2  & \textbf{57.0} \\
    \bottomrule
    \end{tabular}%
  \label{tab5}%
\end{table}%

\subsection{Generalization to different KDs}
\label{4.4}
To further explore the universality of the proposed "\textbf{UET}" paradigm, we conduct a series of experiments by introducing the knowledge uncertainty into different KD methods. Similarly, we employ the GFL framework for this part, following the configurations outlined in Section \ref{4.2}. Moreover, $N$ was set to 5, with a ratio strategy of \textbf{\textit{b)}} for our uncertainty estimate. Specifically, we conduct the generalization experiments on three advanced feature distillation methods, (MGD \cite{23}, PKD \cite{6}, and FGD \cite{7}), as shown in Table \ref{tab6}. Upon introducing knowledge uncertainty, MGD, PKD, and FGD yielded improvements of 2.8\%, 2.4\%, and 3.8\%, respectively, for the student model, surpassing their corresponding pure versions of KDs. These results confirm the applicability of the proposed paradigm across different feature-based KD situations and also suggest that incorporating knowledge uncertainty can optimize the KD process to a certain extent.
\begin{table}[htbp]
  \centering
  \caption{Generalization to different KDs. * denotes our reproduced results.}
    \begin{tabular}{lccccccc}
    \toprule
    \multicolumn{1}{c}{Method} & Schedule & mAP   & $AP_{50}$  & $AP_{75}$ & $AP_S$  & $AP_M$   & $AP_L$ \\
    \midrule
    GFL-Res101 (T) & 2$\times$, ms    & 44.9  & 63.1  & 49.0  & 28.0  & 49.1  & 57.2 \\
    GFL-Res50 (S) & 1$\times$    & 40.2  & 58.4  & 43.3  & 23.3  & 44.0  & 52.2 \\
    \midrule
    \multicolumn{8}{c}{\textit{Logits-based KDs}} \\
    LD*  \cite{19}  & 1$\times$    & 41.0 (+0.8)  & 58.6  & 44.2  & 23.4  & 45.0  & 53.1 \\
    %\textcolor{red}{\textbf{logit-side+ours}} & 1x    & 40.8 (+0.6) & 58.0 & 44.4 & 24.0 & 44.9 & 51.9 \\
    \rowcolor[rgb]{ .906,  .902,  .902} LD+ours & 1$\times$    & \textbf{41.2 (+1.0)} & 58.7 & 44.5 & 24.4 & 45.1 & 52.9 \\
    \midrule
    \multicolumn{8}{c}{\textit{Feature-based KDs}} \\
    MGD* \cite{23}   & 1$\times$    & 42.1 (+1.9) & 60.3  & 45.8  & 24.4  & 46.2  & 54.7 \\
    \rowcolor[rgb]{ .906,  .902,  .902} MGD+Ours & 1$\times$    & \textbf{43.0 (+2.8)} & 61.3  & 46.4  & 26.1  & 47.1  & 55.7 \\
    PKD* \cite{6}  & 1$\times$    & 42.5 (+2.3) & 60.9  & 46.0  & 24.2  & 46.7  & 55.9 \\
    \rowcolor[rgb]{ .906,  .902,  .902} PKD+Ours & 1$\times$    & \textbf{42.6 (+2.4)} & 60.6  & 46.1  & 23.9  & 46.7  & 55.8 \\
    FGD* \cite{7}  & 1$\times$    & 43.4 (+3.2) & 61.7  & 47.0  & 26.2  & 47.4  & 56.4 \\
    \rowcolor[rgb]{ .906,  .902,  .902} FGD+Ours & 1$\times$    & \textbf{44.0 (+3.8)} & 62.2  & 47.7  & 26.8  & 48.4  & 57.0 \\

    \bottomrule
    \end{tabular}%
  \label{tab6}%
\end{table}%

\textbf{Extended to logits-based distillation.} In Section \ref{4.4}, we reveal that introducing knowledge uncertainty into the feature-based distillation process would contribute to student learning. A pertinent question arises: Could the knowledge uncertainty also be applied to logits-based distillation? Compared to feature-based distillation, logit-based distillation directly transfers the output results of the teacher detector, such as classification scores and detection positions. LD \cite{19} is one of the representative logits-based distillation methods, which designs a valuable localization region to distill the student detector in a separate distillation region manner. Following the uncertainty estimation manner in feature-based distillation, we introduce it in the output features of the teacher detector's FPN network and then feed them into the head to obtain predicted logits, as shown in Table \ref{tab6}. As expected, after introducing our method, the detector's mAP reached 41.2\%, compared to the original LD's mAP of 41.0\%. This result further illustrates the importance of introducing knowledge uncertainty for knowledge distillation in object detection.

\subsection{Discussions and analysis}
\textbf{Extended to lightweight detectors.}
Expanding to lightweight detectors presents additional challenges in KD, as the knowledge gap between teachers and students tends to be larger. Furthermore, due to the limitations in learning capacity, student models may not fully comprehend the knowledge of the teacher model. This implies that knowledge incomprehensible to these student models may affect their learning process as noise. Therefore, introducing knowledge uncertainty in the teacher model may be more crucial in the KD process for lightweight detectors. To further explore this, we conduct experiments using the GFL framework, and the results are presented in Table \ref{tab8}. Compared with the setting of Section \ref{4.1}, we only alternate the ResNet50 with a lightweight backbone in the student detector. As hypothesized, introducing knowledge uncertainty is crucial in lightweight detectors, leading to significant improvements. When the student model with ResNet34 as the backbone, introducing knowledge uncertainty results in a detector achieving 41.9\% mAP, which is 2.2\% higher than the 39.7\% obtained with pure FGD. We analyze the impact of introducing knowledge uncertainty on the model's convergence in lightweight detectors in Section \ref{A4}.
\begin{table}[htbp]
  \centering
  \caption{Quantitative results for lightweight detectors.}
    \begin{tabular}{ccccccccc}
    \toprule
    Student &  FGD   & Ours  & mAP   & $AP_{50}$  & $AP_{75}$ & $AP_S$  & $AP_M$   & $AP_L$ \\
    \midrule
    %\rowcolor[rgb]{ .906,  .902,  .902}  Res101 (T) & &     & 44.9  & 63.1  & 49.0  & 28.0  & 49.1  & 57.2 \\
    \multirow{3}[2]{*}{Res18} &       &       & 35.8  & 53.1  & 38.2  & 18.9  & 38.9  & 47.9 \\
          & $\checkmark$     &       & 33.3 (-2.5) & 49.2  & 36.0 & 20.3  & 36.1 & 42.7\\
          & $\checkmark$   & $\checkmark$      & \textbf{37.9 (+2.1) } & 54.9  & 41.0  & 21.9  & 41.5  & 49.3 \\
    \midrule
    \multirow{3}[2]{*}{Res34} &       &       & 38.9  & 56.6  & 42.2  & 21.5  & 42.8  & 51.4 \\
          & $\checkmark$      &       & 39.7 (+0.8)  & 57.1  & 43.0  & 23.0  & 43.7  & 51.2 \\
          & $\checkmark$      & $\checkmark$      & \textbf{41.9 (+3.0) }& 59.6  & 45.3  & 24.3  & 46.0  & 54.4 \\
    % \midrule
    % \multirow{3}[2]{*}{Res50} &       &       & 40.2  & 58.4  & 43.3  & 23.3  & 44.0  & 52.2 \\
    %       & $\checkmark$      &       & 43.4  & 61.7  & 47.0  & 26.2  & 47.4  & 56.4 \\
    %       & $\checkmark$      & $\checkmark$      & 44.1  & 62.3  & 47.8  & 26.6  & 48.2  & 56.9 \\
    \bottomrule
    \end{tabular}%
  \label{tab8}%
\end{table}%

\textbf{Extended to the backbone of the different types.}
We conduct experiments on the RetinaNet detector with heterogeneous and homogeneous backbones, and more details are listed in Section \ref{A2}.

\textbf{Visualization of detection results.}
We validate the effectiveness of our proposed method through visualization evidence, and more details are listed in Section \ref{A5}.

% \begin{figure}
%   \centering
%   \includegraphics[scale=0.5]{2.pdf}
%   \caption{Convergence Analysis}
%   \label{method}
% \end{figure}

% \begin{figure}
%   \centering
%   \includegraphics[scale=0.5]{3.pdf}
%   \caption{Convergence Analysis}
%   \label{method}
% \end{figure}

\section{Conclusions}
In this paper, we investigate the significance of introducing knowledge uncertainty from teacher models in object detection distillation and its impact on the learning process of student detectors. Building upon this, we propose a novel \textbf{UET} general paradigm for feature distillation, aimed at facilitating the acquisition of latent knowledge by student models, while easily being adaptable to other distillation methods. Additionally, we present a simple yet effective uncertainty estimation approach by integrating MC dropout, which seamlessly introduces uncertainty knowledge at minimal computational cost. Extensive experiments validate the effectiveness of following the \textbf{UET} paradigm across various types of KDs, detectors, and backbones. In summary, we demonstrate that introducing teacher knowledge uncertainty enhances the learning capabilities of student models in the KD process.

\bibliographystyle{ieeetr}
\bibliography{neurips_2024}

%%%%%%%%%%%%%%%%%%%%%%%%%%%%%%%%%%%%%%%%%%%%%%%%%%%%%%%%%%%%
\newpage
\appendix
\section{Appendix / supplemental material}
\subsection{Limitations.}
\label{A1}
In this paper, we utilize MC dropout to estimate the uncertainty in the teacher's knowledge. In reality, more advanced uncertainty estimation methods could potentially further enhance the impact of the \textbf{UET} paradigm in knowledge distillation. Additionally, we incorporate estimated uncertainty into the original teacher knowledge using the simple structure of residual learning. Employing more effective integration techniques may yield even better performance. However, in this work, the main insight is that we should seriously consider the uncertainty of teachers' knowledge, and we also provide a simple solution for the KD process with knowledge uncertainty. 

\subsection{Convergence Analysis.}
\label{A4}
\begin{figure}[htbp]
  \centering
  \begin{subfigure}{0.5\textwidth}
    \centering
    \includegraphics[scale=0.5]{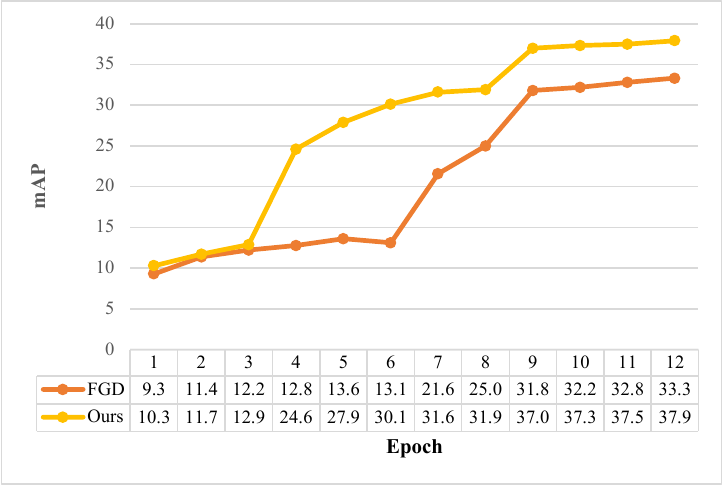}
    \caption{The lightweight detector with ResNet18.}
    \label{fig:a}
  \end{subfigure}%
  \begin{subfigure}{0.5\textwidth}
    \centering
    \includegraphics[scale=0.535]{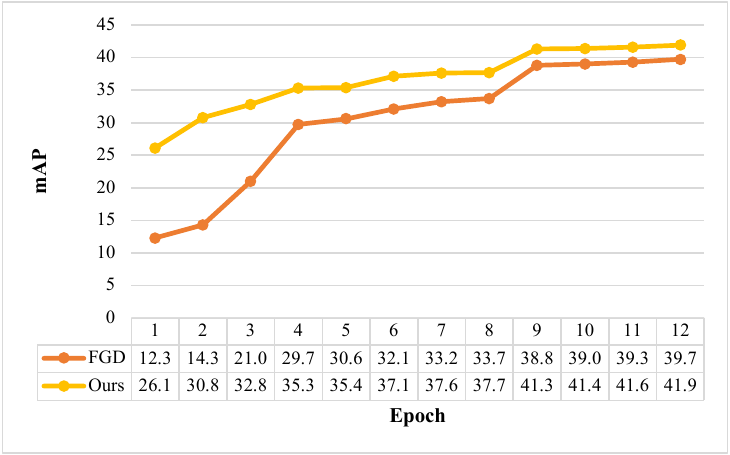}
    \caption{The lightweight detector with ResNet34.}
    \label{fig:b}
  \end{subfigure}
  \caption{Convergence analysis for the lightweight detectors}
  \label{fig2}
\end{figure}

We also analyze the impact of introducing knowledge uncertainty on the model's convergence in lightweight detectors. As illustrated in Figure \ref{fig2}, when we introduce knowledge uncertainty during training, the model's convergence speed significantly improved. Particularly, with ResNet34 as the backbone, the student model following our paradigm even outperformed the baseline's performance in the third epoch. This indicates that introducing knowledge uncertainty in KD detection not only enhances student learning ability but also accelerates the convergence speed.

\subsection{Experiments for detectors with heterogeneous and homogeneous backbone.}
\label{A2}
We conduct experiments on the RetinaNet detector with both heterogeneous and homogeneous backbones, and the results are presented in Table \ref{tab9}. Unsurprisingly, regardless of the backbone configuration being heterogeneous or homogeneous, the FGD method demonstrates stronger knowledge transfer capabilities after following the UET paradigm. For instance, under the heterogeneous backbone setting of SwinT \cite{35}, the student detector achieves 37.7\% mAP after introducing knowledge uncertainty, surpassing the 37.4\% obtained with pure FGD.
\begin{table}[htbp]
  \centering
  \caption{Experiments for detectors with Heterogeneous and homogeneous backbone.}
    \begin{tabular}{cccccccc}
    \toprule
    Methods & Schedule & mAP   & $AP_{50}$  & $AP_{75}$ & $AP_S$  & $AP_M$   & $AP_L$ \\
    \midrule
     \rowcolor[rgb]{ .906,  .902,  .902}SwinT \cite{35} (T) & 1$\times$  & 37.3  & 57.5  & 39.9  & 22.7  & 41.0  & 49.6 \\
     \rowcolor[rgb]{ .906,  .902,  .902}Res50 \cite{13} (S) & 1$\times$    & 36.5  & 55.4  & 39.1  & 20.4  & 40.3  & 48.1 \\
    FGD \cite{7}  & 1$\times$    & 37.4 (+0.9) & 56.8  & 39.9  & 22.6  & 41.3  & 48.7 \\
    FGD+Ours & 1$\times$    & \textbf{37.7 (+1.2)} & 57.2  & 40.2  & 21.9  & 41.7  & 59.0 \\
    \midrule
      \rowcolor[rgb]{ .906,  .902,  .902}Res50 \cite{13} (T) & 1$\times$    & 36.5  & 55.4  & 39.1  & 20.4  & 40.3  & 48.1 \\
      \rowcolor[rgb]{ .906,  .902,  .902}Res50 \cite{13} (S) & 1$\times$    & 36.5  & 55.4  & 39.1  & 20.4  & 40.3  & 48.1 \\
    FGD  \cite{7} & 1$\times$    & 37.4 (+0.9) & 56.7  & 39.7  & 20.6  & 40.9  & 49.0 \\
    FGD+Ours & 1$\times$    & \textbf{37.9 (+1.4)} & 57.3  & 40.4  & 21.2  & 41.6  & 50.1 \\
    \bottomrule
    \end{tabular}%
  \label{tab9}%
\end{table}%

\subsection{The pseudo-code of our UET paradigm}
In comparison to the \textbf{ET} paradigm, we introduce the \textbf{UET} paradigm by incorporating knowledge uncertainty. Building upon the \textbf{ET} paradigm, we estimate the uncertainty in the teacher detector's knowledge using Eq. \ref{5} and then perform knowledge transfer according to Eq. \ref{6}. Existing feature distillation methods can transform from the \textbf{ET} to the \textbf{UET} paradigm with minimal additional computational overhead (only requires a change in the \textcolor{blue}{\textbf{high light}} parts of code).

\begin{algorithm}
  \caption{UET Paradigm}
  \begin{algorithmic}[1]
    \State \textbf{Require: } Training data $x_{i\{i=1,...,n\}}$, 
    
    FPN network of student detector $S_{det}$, 
    
    FPN network of teacher detector $T_{det}$,    
    \State Uniformly sample a minibatch of training data $B^{(t)}$  
    \For{ $x_i \in B^{(t)}$}
        \State $F^{T} = T_{det}(x_i)$;
        \State $F^{S} = S_{det}(x_i)$;
        \Procedure{UET\_Paradigm}{$F^{T}, F^{S}$}
        
        \State \textcolor{blue}{Estimate Uncertainty:} 
        $\textcolor{blue}{U_K = \frac{1}{N} \sum_{i=1}^{N} Dropout_i (F^{(T)}(x))}$
        
        \State  KD with Knowledge uncertainty: 
        
        $\arg\min_{\theta}  \mathcal{L}_{\text{KD}} (\theta) = d_\textbf{T}(f_\textbf{E}(\textcolor{blue}{U_K + F^{(T)}(x)}), f_\textbf{E}(g(F^{(S)}(x, \theta))))$,
        \EndProcedure
    \EndFor
  \end{algorithmic}
\label{A3}
\end{algorithm}

\subsection{Visualization of detection results}
\label{A5}
\begin{figure}[htbp]
  \centering
  \includegraphics[scale=0.225]{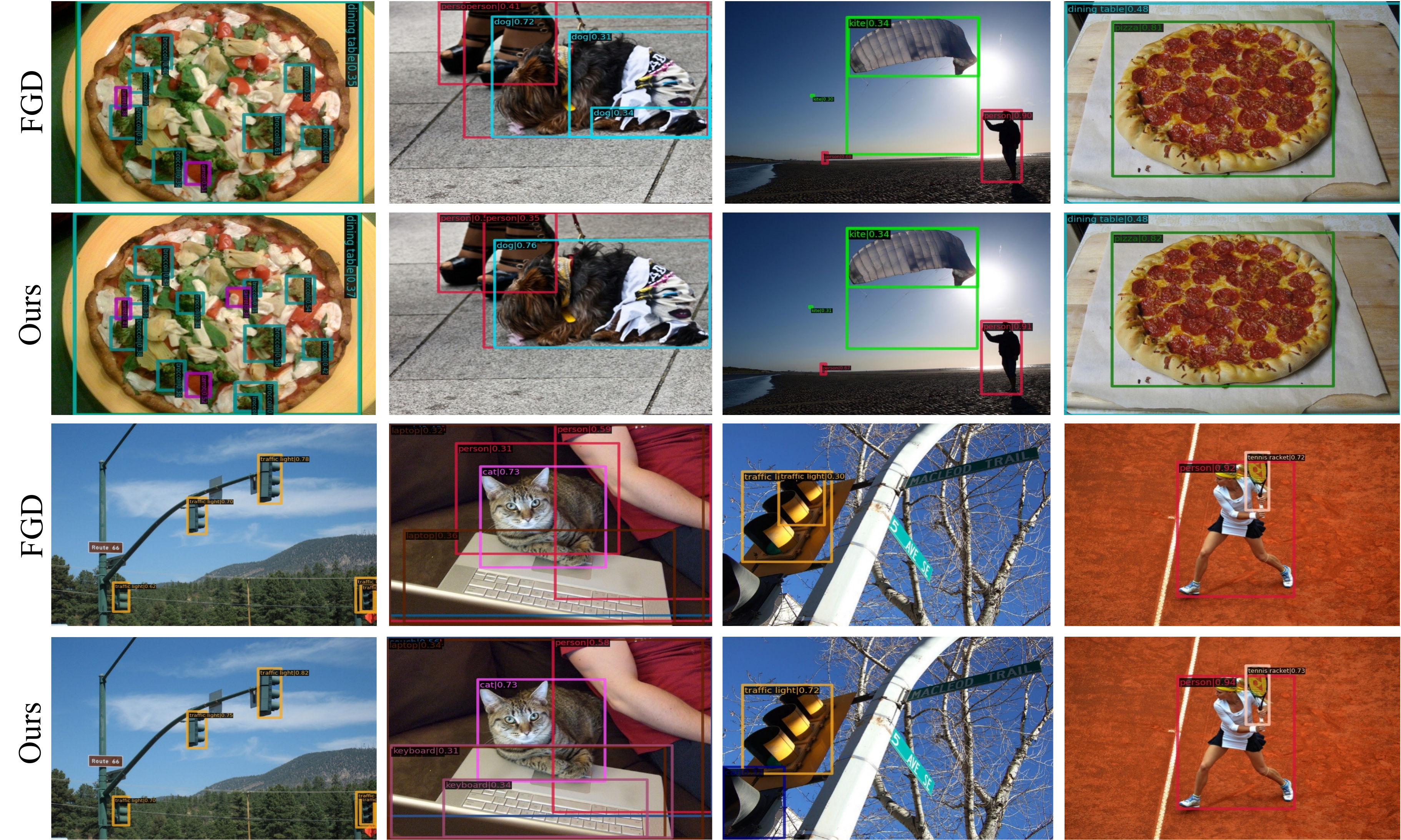}
  \caption{Visualization of detection results of Our method.}
  \label{vs}
\end{figure}

We provide visual evidence to validate the effectiveness of our proposed method by presenting detection results from the val2017 set of MS COCO \cite{12}. Figure \ref{vs} shows that following our \textbf{UET} paradigm allows FGD to outperform the original FGD in detecting more high-quality bounding boxes. This suggests that our method effectively enhances the learning potential of student detectors during the KD process.

\end{document}